\documentclass[journal]{IEEEtran}
\renewcommand{\baselinestretch}{0.98}

\usepackage{cite}
\usepackage{amsmath,amssymb,amsfonts}
\usepackage{graphicx}
\usepackage{textcomp}
\usepackage{xcolor}
\usepackage{url}
\usepackage[nobiblatex]{xurl}
\usepackage{multirow}
\usepackage{longtable}
\usepackage{tabularx}
\usepackage{booktabs}
\usepackage{array}
\usepackage{placeins}
\usepackage{enumitem}
\usepackage{algorithm}
\usepackage{algpseudocode}

\def\BibTeX{{\rm B\kern-.05em{\sc i\kern-.025em b}\kern-.08em
    T\kern-.1667em\lower.7ex\hbox{E}\kern-.125emX}}

\setlist{topsep=2pt,itemsep=1pt,parsep=0pt,partopsep=0pt}

\begin{document}

\title{AI-Enhanced Spatial Cellular Traffic Demand Prediction with Contextual Clustering and Error Correction for 5G/6G Planning}

\author{Mohamad~Alkadamani, Colin~Brown, and Halim~Yanikomeroglu%
\thanks{M. Alkadamani is with Innovation, Science and Economic Development Canada (ISED) and Carleton University, Ottawa, Canada (e-mail: mohamad.alkadamani@ised-isde.gc.ca).}%
\thanks{C. Brown is with the Communications Research Centre (CRC), Ottawa, Ontario, Canada.}%
\thanks{H. Yanikomeroglu is with Carleton University, Ottawa, Ontario, Canada.}%
}

\maketitle

\begin{abstract}
Accurate spatial prediction of cellular traffic demand is essential for 5G NR capacity planning, network densification, and data-driven 6G planning. Although machine learning can fuse heterogeneous geospatial and socio-economic layers to estimate fine-grained demand maps, spatial autocorrelation can cause neighborhood leakage under naive train/test splits, inflating accuracy and weakening planning reliability. This paper presents an AI-driven framework that reduces leakage and improves spatial generalization via a context-aware two-stage splitting strategy with residual spatial error correction. Experiments using crowdsourced usage indicators across five major Canadian cities show consistent mean absolute error (MAE) reductions relative to location-only clustering, supporting more reliable bandwidth provisioning and evidence-based spectrum planning and sharing assessments.
\end{abstract}


\section{Introduction}\label{sec:introduction}
A key capability for AI-enabled wireless networks and cognitive spectrum management is characterizing how cellular traffic demand varies across space, particularly in dense urban regions. For 5G and beyond, spatial demand hotspots influence carrier bandwidth selection, small-cell deployment, capacity expansion, and the feasibility of spectrum access mechanisms. Spatial demand prediction can also help regulators and planners identify regions at risk of under- or over-provisioning and screen spectrum-sharing feasibility in longer-horizon planning.

Estimating cellular traffic demand, especially for commercial mobile applications, is complex and shaped by inter-related factors such as technological evolution, regulations, usage trends, and socio-economic conditions. Traditionally, planning studies relied on expert assessments in industry whitepapers~\cite{GSMA_2021} or simplified models~\cite{ITU_M1768}. Increasing system complexity and data availability have accelerated interest in data-driven planning and resource-allocation methods~\cite{doke2024,FCC_data_driven,brown2024}. These approaches, including ML and advanced analytics, can infer spatial demand distributions and support planning decisions via predictive traffic-demand maps~\cite{ParekhFNWF2023}.

Data-driven spatial traffic-demand prediction faces challenges specific to geospatial wireless data, including heterogeneous feature resolutions, high dimensionality, and the need to handle spatial autocorrelation to avoid biased evaluation. A central concern is \emph{spatial information leakage}: nearby samples are statistically dependent, so naive train/test splitting can yield over-optimistic accuracy and misleading generalization claims.

Prior work on traffic-demand proxy modeling provides a starting point. In~\cite{ParekhPimrc2023}, diverse geospatial inputs were studied alongside a demand proxy, and location-based clustering was used to improve train/test independence;~\cite{ParekhFNWF2023} advanced the proxy and interpretability aspects. Spatio-temporal traffic prediction has also been studied using recurrent neural networks with geo-referenced traffic measurements to capture temporal dynamics and geographic variability~\cite{Qiu2018SpatioTemporal}. From the leakage-mitigation perspective, spatial cross-validation techniques are surveyed in~\cite{gao2023,SALAZAR2022}, and related fields propose cluster-based and hybrid splitting strategies~\cite{feng2023,wang2023}. However, many clustering-based splits remain \emph{context-blind}, relying mainly on location (or generic clustering) without explicitly enforcing representativeness in land-use and functional context; consequently, dependence can still leak across folds and evaluation may remain optimistic.

This paper proposes a context-aware framework for spatial cellular traffic-demand prediction that improves spatial generalization under spatial autocorrelation and supports reliable 5G/6G planning. Contributions include a context-aware two-stage splitting strategy that combines spatial clustering with land-use/context clustering to form representative folds and reduce leakage relative to location-only clustering, and a five-city Canadian evaluation with planning-oriented mappings that translate prediction error into bandwidth-dimensioning error and congestion-risk screening for carrier bandwidth and spectrum-provisioning assessments.

The remainder of the paper is organized as follows. Section~\ref{sec:data_driven_modeling} describes the data model and feature mapping. Section~\ref{sec:spatial_dependency} characterizes spatial dependency using Moran’s I. Section~\ref{sec:data_splitting} presents the proposed two-stage splitting strategy. Section~\ref{sec:performance_evaluation} reports performance results, Section~\ref{sec:wireless_link} links errors to 5G/6G planning metrics, and Section~\ref{sec:conclusions} concludes.

\section{Data-driven Spatial Traffic Demand Modeling}\label{sec:data_driven_modeling}
Spatial traffic-demand prediction is formulated as supervised learning over heterogeneous geospatial features by mapping both feature layers and a traffic-demand proxy to a common geographic unit.

\subsection{Grid-cell representation and study areas}
The study region is partitioned into uniform square \textit{grid cells} of approximately $1.5\,\text{km}\times1.5\,\text{km}$. The grid cell is an analysis unit used to align datasets to a common spatial resolution. Five major Canadian cities are considered: Montreal, Vancouver, Greater Toronto, Ottawa, and Calgary. Let $S=\{s_1,\ldots,s_n\}$ denote the set of $n$ grid cells. Each $s_i$ has a feature vector $\mathbf{x}_i\in\mathbb{R}^m$ and a target $y_i$ representing a traffic-demand proxy, with learning objective $\hat{y}_i=f_\theta(\mathbf{x}_i)$.

\subsection{Traffic demand proxy from crowdsourced measurements}
Direct traffic measurements are typically restricted to mobile network operators and are unavailable for public studies. A traffic-demand proxy $y_i$ is therefore derived from crowdsourced mobile usage indicators collected via application-embedded SDKs and aggregated to the grid-cell level. The dataset comprises approximately 15 million measurements across the five cities over one month in 2023.

In the 4G/5G transition regime, the indicators reflect combined activity and represent total traffic intensity rather than generation-specific load. The proxy increases with the number and persistence of observed user connections within a grid cell and, although bytes transferred are not explicitly encoded, captures the dominant busy-hour spatial structure (dense connections imply higher load, while sparse activity implies lower demand). After filtering and aggregation, each grid cell $s_i$ is assigned $y_i$ paired with $\mathbf{x}_i$ for supervised learning.

\subsection{Feature layers and grid-cell mapping}
Input features originate from heterogeneous sources and are standardized to the grid-cell representation. Feature layers may be defined on administrative units (e.g., census subdivisions and dissemination areas), as points (e.g., points of interest and businesses), or as polygons (e.g., land-use/land-cover). Geometry-aware mapping assigns each grid cell $s_i$ a consistent feature vector $\mathbf{x}_i$.

Feature layers are temporally aligned to overlap with the same one-month window of the crowdsourced dataset; some layers have coarser temporal granularity than others but still overlap the analysis period. For areal datasets, grid-cell values are obtained by spatial overlay using area-weighted allocation from source polygons to intersecting grid cells. For categorical polygon layers (e.g., land-use), the dominant class within each grid cell is used; point-based layers are mapped via counts or density estimates (normalized by cell area).

\begin{figure}[t]
    \centering
    \includegraphics[width=\columnwidth]{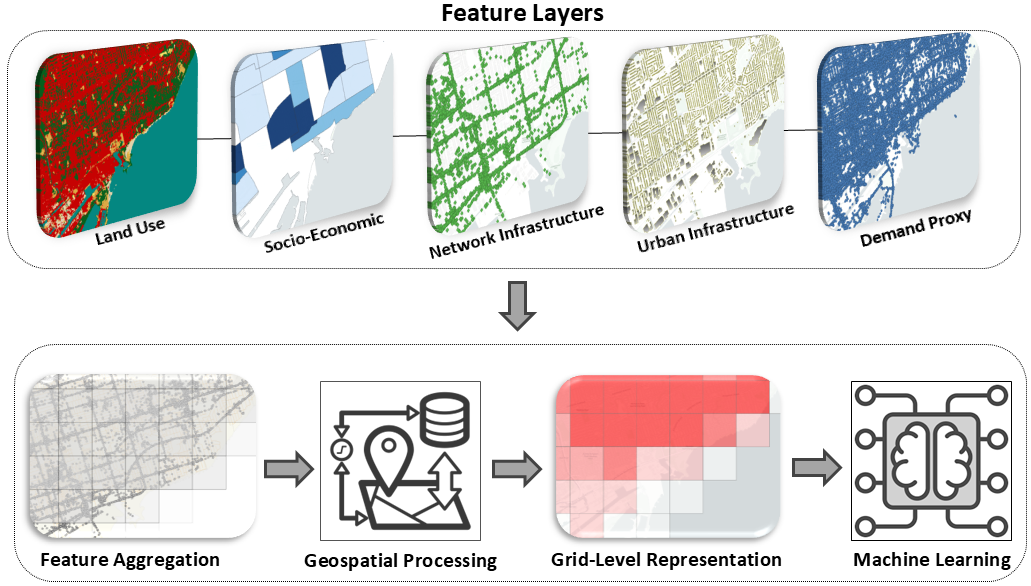}
    \caption{Modeling pipeline.}
    \label{fig:spectrum_demand_pipeline}
\end{figure}

Mapped features include socio-economic variables (population density and related census indicators), urban infrastructure (counts/densities of businesses, roads, and points of interest), land-use type (land-use/land-cover attributes), and network infrastructure (an indicator of cellular infrastructure presence). All layers are aligned to the grid and combined into $\mathbf{x}_i$, yielding $\{(\mathbf{x}_i,y_i)\}_{i=1}^{n}$. Figure~\ref{fig:spectrum_demand_pipeline} summarizes the pipeline.

\section{Spatial Dependency Characterization}\label{sec:spatial_dependency}
Urban traffic-demand maps exhibit spatial dependence: neighboring areas often share land use, socio-economic conditions, and mobility-driven usage, producing correlated demand values. Quantifying spatial dependence (i) motivates leakage-aware splitting and (ii) informs the spatial scale required to separate neighborhoods and obtain realistic generalization estimates.

\subsection{Global Moran's I for spatial autocorrelation}
Global Moran's I quantifies city-wide spatial autocorrelation in $y_i$~\cite{Morans}. For $N$ grid cells,
\begin{equation}
I = \frac{N}{W}\,\frac{\sum_{i=1}^{N}\sum_{j=1}^{N} w_{ij}(y_i - \bar{y})(y_j - \bar{y})}{\sum_{i=1}^{N}(y_i - \bar{y})^2},
\end{equation}
where $\bar{y}$ is the mean demand proxy, $w_{ij}$ is a spatial weight, and $W=\sum_{i=1}^{N}\sum_{j=1}^{N} w_{ij}$.

\begin{figure}[t]
    \centering
    \includegraphics[width=\columnwidth]{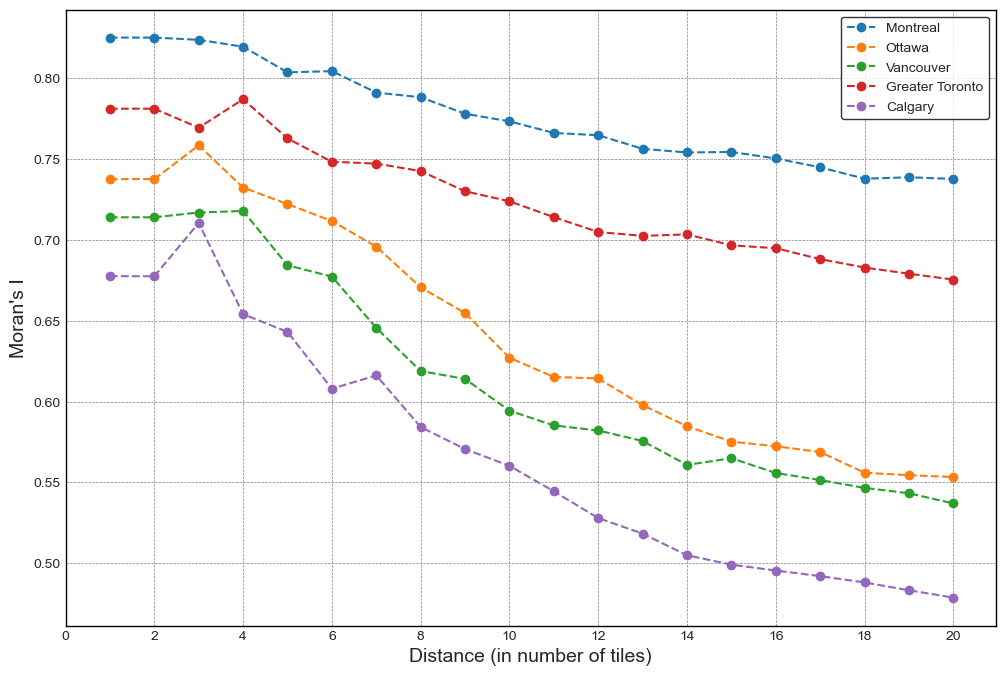}
    \caption{Moran's I versus distance (grid cells).}
    \label{fig:spatial_dependency_distance}
\end{figure}

A distance-threshold neighborhood is used with $w_{ij}=1$ if centroid distance $d_{ij}\le d_{\text{threshold}}$ and $w_{ij}=0$ otherwise; varying $d_{\text{threshold}}$ reveals the correlation range. Figure~\ref{fig:spatial_dependency_distance} plots $I$ versus distance (in grid cells) to identify the dominant correlation range that split boundaries should exceed to reduce leakage; inter-city differences further indicate that a single fixed split radius can be suboptimal.

\subsection{Local Moran's I for spatial clusters and outliers}
Local Moran's I localizes spatial clusters and outliers. For grid cell $s_i$,
\begin{equation}
I_i = (y_i - \bar{y}) \sum_{j=1}^{N} w_{ij} (y_j - \bar{y}),
\end{equation}
where $w_{ij}$ defines the neighborhood.

\begin{figure}[t]
    \centering
    \includegraphics[width=\columnwidth]{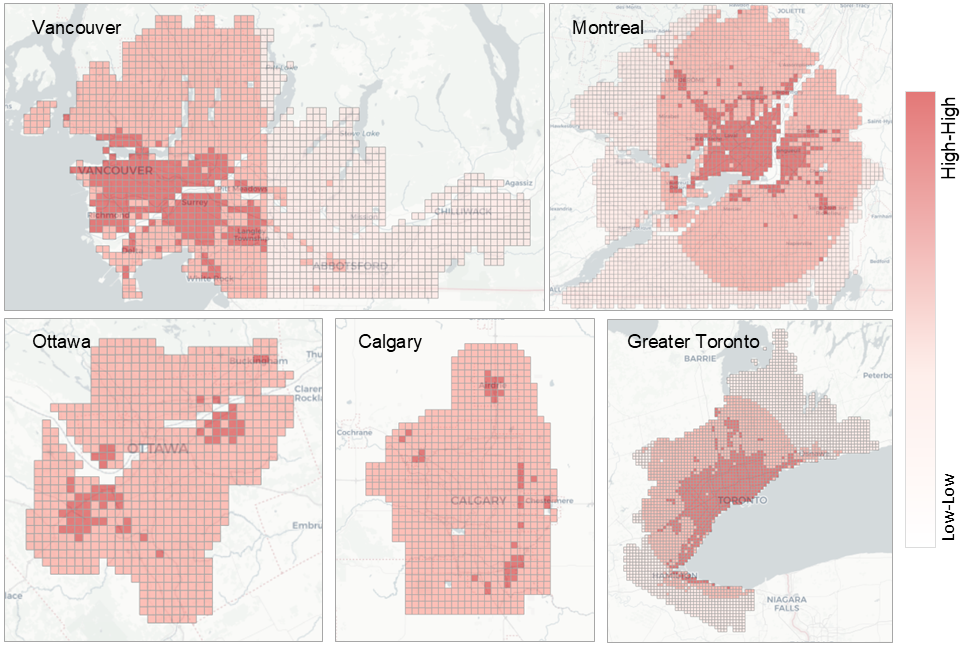}
    \caption{Local Moran's I clusters across five Canadian cities.}
    \label{fig:local_morans_clusters}
\end{figure}

Local Moran’s I yields four spatial association types: High-High (HH), high $y_i$ with high neighbors; Low-Low (LL), low $y_i$ with low neighbors; High-Low (HL), high $y_i$ with low neighbors; and Low-High (LH), low $y_i$ with high neighbors. Figure~\ref{fig:local_morans_clusters} visualizes these associations, with dark red indicating HH clusters (persistent hotspots) and light red indicating LL clusters (persistent low-demand regions). Transitional areas often exhibit HL/LH behavior and can be challenging for generalization when neighbors are split across folds.

\section{Two-Stage Data Splitting and Spatial Error Correction}\label{sec:data_splitting}
\begin{figure*}
\includegraphics[width=\linewidth]{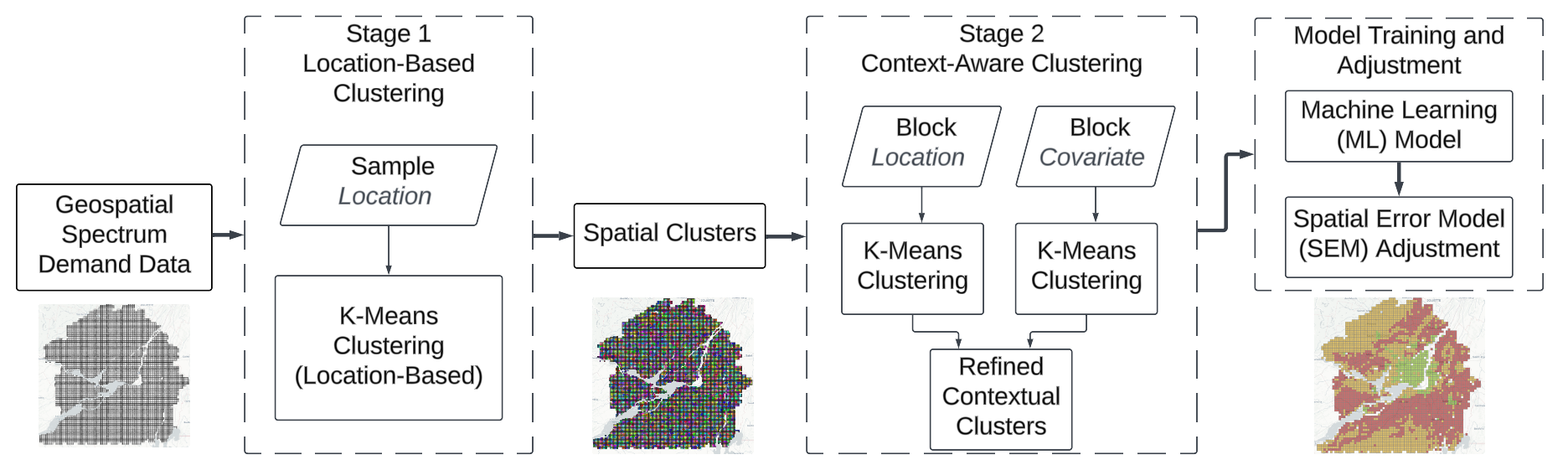}
\caption{Two-stage clustering and spatial error correction framework.}
\label{fig:two_stage_cluster}
\end{figure*}

Spatial autocorrelation makes random splitting unreliable because adjacent grid cells can share near-duplicate context, inflating validation accuracy when neighbors fall in different folds. Location-only clustering reduces leakage but can produce context-imbalanced folds (e.g., commercial versus residential); the proposed two-stage procedure enforces spatial separation and context diversity and is followed by spatial error correction  as shown in Fig.~\ref{fig:two_stage_cluster}.

\subsection{Stage 1: spatial clustering for leakage reduction}
Stage~1 partitions the study area into spatially cohesive blocks by applying k-Means to grid-cell centroids
\begin{equation}
\min_{\{C_i\}_{i=1}^k} \sum_{i=1}^{k} \sum_{s_j \in C_i} \| \mathbf{p}_j - \boldsymbol{\mu}_i \|^2,
\end{equation}
where $\mathbf{p}_j$ is the coordinate vector of $s_j$, $C_i$ is the $i$th spatial cluster, and $\boldsymbol{\mu}_i$ is its centroid. The parameter $k$ controls cluster diameter and the extent to which correlated neighbors are separated across folds; Fig.~\ref{fig:spatial_dependency_distance} guides selection by indicating the dominant correlation range (up to $r$ grid cells) that most train/validation boundaries should exceed.

\subsection{Stage 2: context-aware refinement within spatial clusters}
Stage~2 refines each spatial cluster using context features (e.g., land-use/land-cover and related attributes) to improve representativeness in feature space and avoid folds dominated by a single context type. A normalized dissimilarity between $s_a$ and $s_b$ within the same spatial cluster is
\begin{equation}
d(s_a, s_b) = \sum_{\ell=1}^{m} \frac{|x_{a\ell} - x_{b\ell}|}{\sigma_\ell},
\end{equation}
where $\sigma_\ell$ is the standard deviation of feature $\ell$. Categorical land-use can be represented via one-hot encoding so that dissimilarity reflects context changes.

\begin{figure}[t]
    \centering
    \includegraphics[width=\columnwidth]{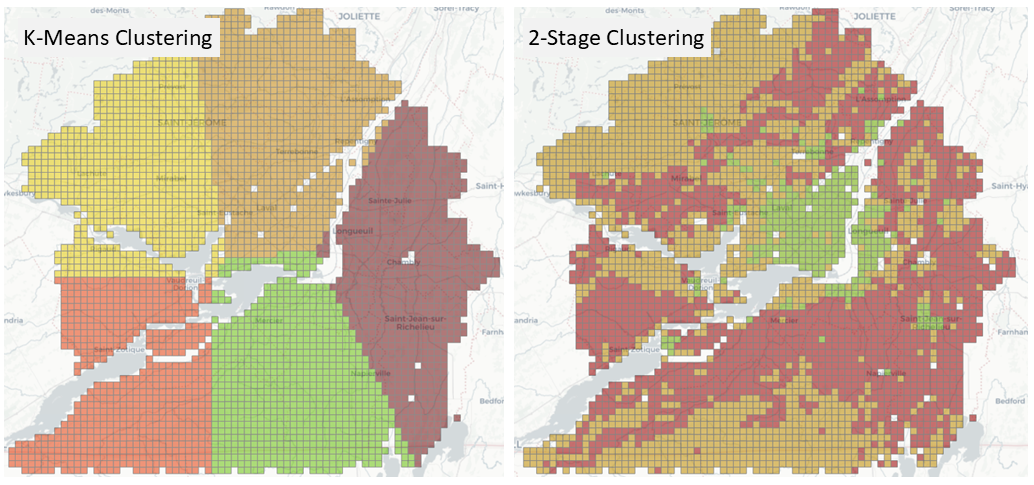}
    \caption{Comparison of clustering techniques for Montreal.}
    \label{fig:comparison_cluster}
\end{figure}

Stage 2 produces sub-clusters that remain spatially contained within Stage-1 blocks and are more homogeneous in context. Sub-clusters are assigned to folds to preserve Stage-1 spatial separation while ensuring each fold contains a mixture of contexts. Figure~\ref{fig:comparison_cluster} illustrates the effect in Montreal. As shown, the location-only k-Means partition uses five folds, whereas the two-stage procedure yields three folds because refinement and fold construction merge sub-clusters into the smallest set of folds that remains spatially separated and sufficiently distinct in land-use context.

\subsection{Learning model and spatial error correction}
XGBoost is used as the base predictor due to strong performance on structured tabular features and nonlinear interactions. Let $\hat{y}_i$ denote the prediction and $e_i=y_i-\hat{y}_i$ the residual. Even under leakage-reduced splitting, residuals can remain spatially correlated due to unmodeled neighborhood effects and latent variables, creating geographically coherent bias. A Spatial Error Model (SEM) is applied to the residual process
\begin{equation}
\mathbf{y} = X\boldsymbol{\beta} + \boldsymbol{\epsilon}, \qquad
\boldsymbol{\epsilon} = \lambda W\boldsymbol{\epsilon} + \mathbf{u},
\end{equation}
where $W$ is a spatial weights matrix (distance threshold or $k$-nearest neighbors), $\lambda$ captures residual spatial dependence, and $\mathbf{u}$ is i.i.d.\ noise. The SEM represents residuals as a spatially filtered process.

SEM refinement is implemented as post-processing: XGBoost is trained on two-stage folds, residuals are computed on training data, $\lambda$ is estimated given $W$, and the spatial filter is applied to correct predictions. A regularized objective that penalizes spatially structured residuals is
\begin{equation}
\mathcal{L}(\theta, \lambda) =
\frac{1}{N}\sum_{i=1}^{N}(\hat{y}_i - y_i)^2
+ \alpha \|\theta\|_2^2
+ \beta \|(I-\lambda W)^{-1}\boldsymbol{\epsilon}\|_2^2,
\end{equation}
where $\theta$ denotes XGBoost parameters. The third term discourages residual autocorrelation, improving robustness in unseen geographic regions.

\section{Performance Evaluation}\label{sec:performance_evaluation}
This section evaluates predictive performance and spatial generalization under the proposed splitting strategy using MAE and $R^2$, with learning curves assessing leakage-driven overfitting under spatially structured validation.

\subsection{Cross-city evaluation}
Table~\ref{tab:performance} reports two evaluations. A leave-one-city-out configuration assesses transferability by holding each city out as an unseen test set while training on the remaining cities. An ``All Cities'' configuration performs splits across the pooled dataset to reflect within-distribution performance.

Across both configurations, two-stage splitting consistently reduces MAE relative to location-only clustering, indicating that context-aware folds better reflect deployment heterogeneity. SEM provides additional MAE reductions, consistent with residual spatial structure not fully captured by the base predictor. Although $R^2$ gains are smaller, the systematic MAE reduction improves absolute demand estimates, which is the critical quantity for downstream planning.

\begin{table}[h]
    \centering
    \caption{Performance Results Across Five Canadian Cities}
    \label{tab:performance}
    \scriptsize
    \begin{tabular}{|p{1.4cm}|p{1.3cm}|p{1.3cm}|p{1.3cm}|p{1.4cm}|}
        \hline
        \multirow{2}{*}{\textbf{City}} & \multicolumn{3}{c|}{\textbf{Mean Absolute Error (MAE)}} & \multirow{2}{*}{\textbf{R\(^2\) Gain (\%)}} \\
        \cline{2-4}
        & \textbf{k-Means} & \textbf{Two-Stage} & \textbf{Two-Stage + SEM} & \\
        \hline
        Toronto & 1532.8 & 1012.3 & 845.2 & 3.85 \\
        Montreal & 1621.5 & 1123.6 & 825.0 & 2.87 \\
        Ottawa & 1475.4 & 987.2 & 808.3 & 3.86 \\
        Vancouver & 1398.6 & 953.5 & 783.4 & 4.88 \\
        Calgary & 1450.7 & 1001.9 & 795.7 & 3.86 \\
        \hline
        \hline
        \textbf{All Cities} & \textbf{1432.7} & \textbf{989.9} & \textbf{806.7} & \textbf{3.66} \\
        \hline
    \end{tabular}
\end{table}

\subsection{Learning curves}
\begin{figure}[h!]
    \centering
    \includegraphics[width=\columnwidth]{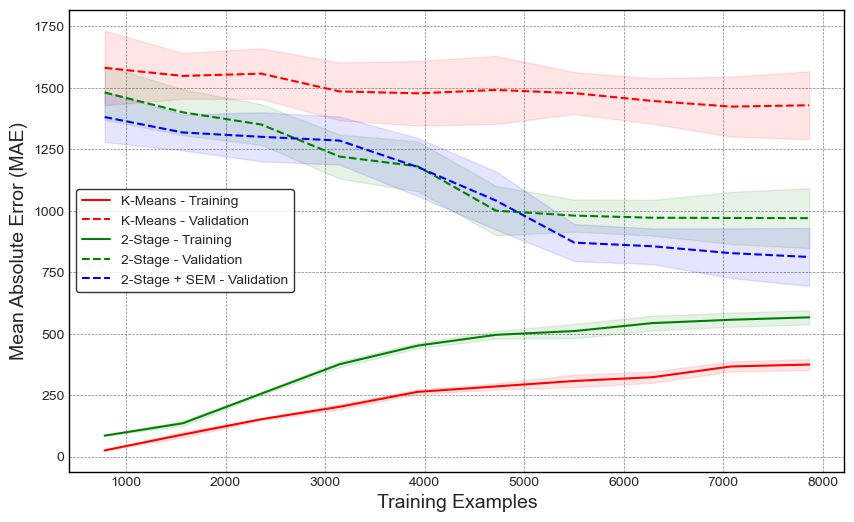}
    \caption{Learning curves comparison for different clustering strategies.}
    \label{fig:learning_curves}
\end{figure}

Figure~\ref{fig:learning_curves} compares learning curves for location-only clustering, two-stage spatial+context splitting, and two-stage splitting with SEM refinement. The training--validation gap indicates overfitting under the chosen split, and shaded bands reflect fold variability.

Location-only clustering shows a large training--validation gap, indicating limited transfer to spatially distinct regions. Two-stage splitting reduces this gap and improves validation MAE, while SEM refinement further lowers validation error by correcting residual spatial dependence.

\section{Wireless Planning Link for 5G/6G}\label{sec:wireless_link}

This section presents a 5G NR mid-band (3.5~GHz) case study mapping the prediction errors in Section~\ref{sec:performance_evaluation} to bandwidth-dimensioning and congestion-risk metrics.

\subsection{Bandwidth dimensioning}
Offered downlink traffic demand (bps) in grid cell $s_i$ is mapped from the proxy $y_i$ as
\begin{equation}\label{eq:proxy_map}
D_i=\kappa y_i,
\end{equation}
where $\kappa$ is the busy-hour traffic per proxy unit; $\kappa=50$~kbps per proxy unit is used across all methods.

Let $\gamma_i$ denote the downlink SINR random variable over the geographic area represented by $s_i$. A planning-level spectral-efficiency abstraction is $\eta(\gamma)=(1-\rho_{\mathrm{oh}})\log_2(1+\gamma)$, where $\eta(\cdot)$ is in bps/Hz and $\rho_{\mathrm{oh}}\in[0,1)$ captures fractional overhead loss. An outage-constrained effective spectral efficiency is defined as $\eta_i^{(\delta)}=Q_{1-\delta}(\eta(\gamma_i))$, where $Q_{1-\delta}(\cdot)$ is the $(1-\delta)$-quantile and $\delta$ is the allowable outage probability. The bandwidth required to serve $s_i$ is approximated by
\begin{equation}\label{eq:breq}
B_{i,\mathrm{req}}^{(\delta)}=\frac{D_i}{\eta_i^{(\delta)}}.
\end{equation}
For a predicted proxy $\hat{y}_i$, the induced per-cell bandwidth-error magnitude follows
$\left|\hat{B}_{i,\mathrm{req}}^{(\delta)}-B_{i,\mathrm{req}}^{(\delta)}\right|
=(\kappa/\eta_i^{(\delta)})\left|\hat{y}_i-y_i\right|$.
Under a constant planning assumption $\eta_i^{(\delta)}=\eta^{(\delta)}$, the mean absolute bandwidth dimensioning error (BDE) is proportional to MAE:
\begin{equation}\label{eq:bde_mae}
\mathrm{BDE}^{(\delta)}=\frac{\kappa}{\eta^{(\delta)}}\cdot \mathrm{MAE}.
\end{equation}

\begin{figure}[t]
    \centering
    \includegraphics[width=\columnwidth]{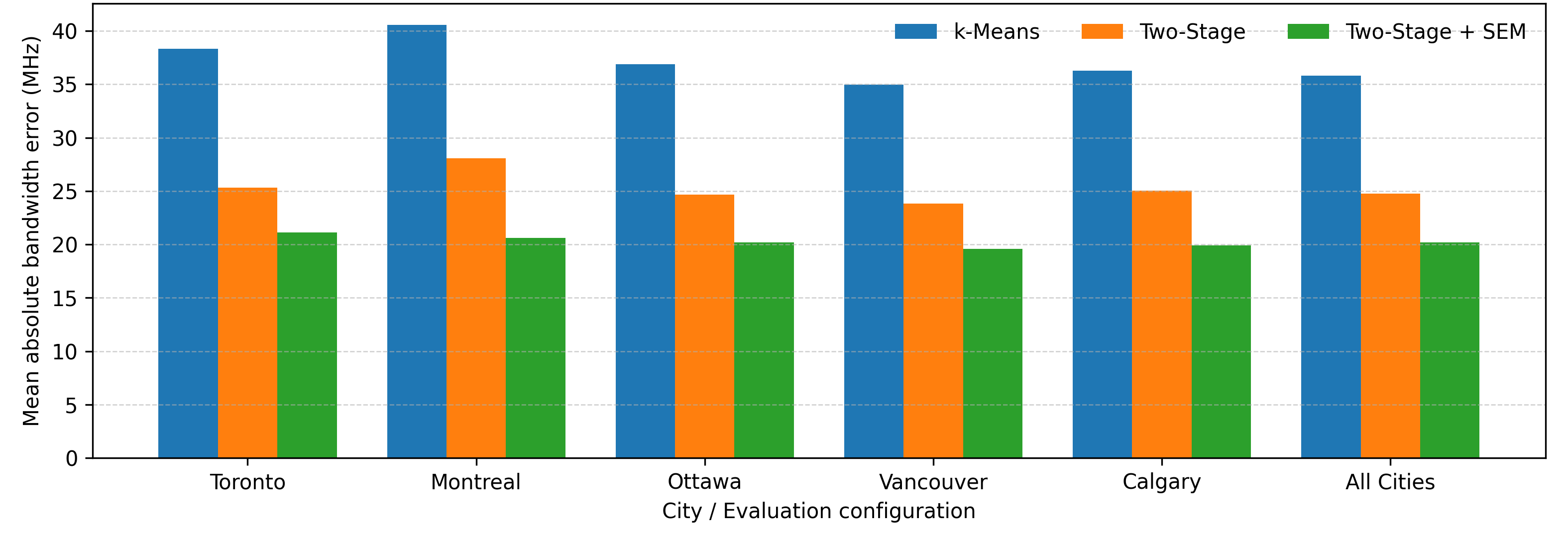}
    \caption{Case study: bandwidth dimensioning error (BDE).}
    \label{fig:bde}
\end{figure}

Table~\ref{tab:bde_sensitivity} reports the resulting ``All Cities'' BDE sensitivity for $\eta^{(\delta)}\in\{2,3,3.5\}$~bps/Hz using the MAE in Table~\ref{tab:performance}, and Fig.~\ref{fig:bde} illustrates the city-wise BDE mapping for the baseline setting $\eta^{(\delta)}=2$~bps/Hz.

\begin{table}[t]
    \centering
    \caption{Sensitivity Table: ``All Cities'' BDE (MHz) for 3.5~GHz.}
    \label{tab:bde_sensitivity}
    \scriptsize
    \begin{tabular}{|c|c|c|c|}
        \hline
        \textbf{$\eta^{(\delta)}$ (bps/Hz)} & \textbf{k-Means} & \textbf{Two-Stage} & \textbf{Two-Stage + SEM} \\
        \hline
        2.0 & 35.8 & 24.7 & 20.2 \\
        3.0 & 23.9 & 16.5 & 13.4 \\
        3.5 & 20.5 & 14.1 & 11.5 \\
        \hline
    \end{tabular}
\end{table}

\subsection{Congestion risk versus candidate carrier bandwidth}
Feasibility screening of candidate carrier bandwidths $B$ (e.g., 40--100~MHz at 3.5~GHz) is captured by the congested fraction
\begin{equation}\label{eq:pcong}
P_{\mathrm{cong}}(B) = \frac{1}{N}\sum_{i=1}^{N}\mathbb{I}\!\left(D_i > B\,\eta_i^{(\delta)}\right),
\end{equation}
where $P_{\mathrm{cong}}(B)$ is the share of grid cells whose offered demand exceeds supported capacity $B\,\eta_i^{(\delta)}$ and $N$ is the number of grid cells in the evaluation set.

\begin{figure}[t]
    \centering
    \includegraphics[width=\columnwidth]{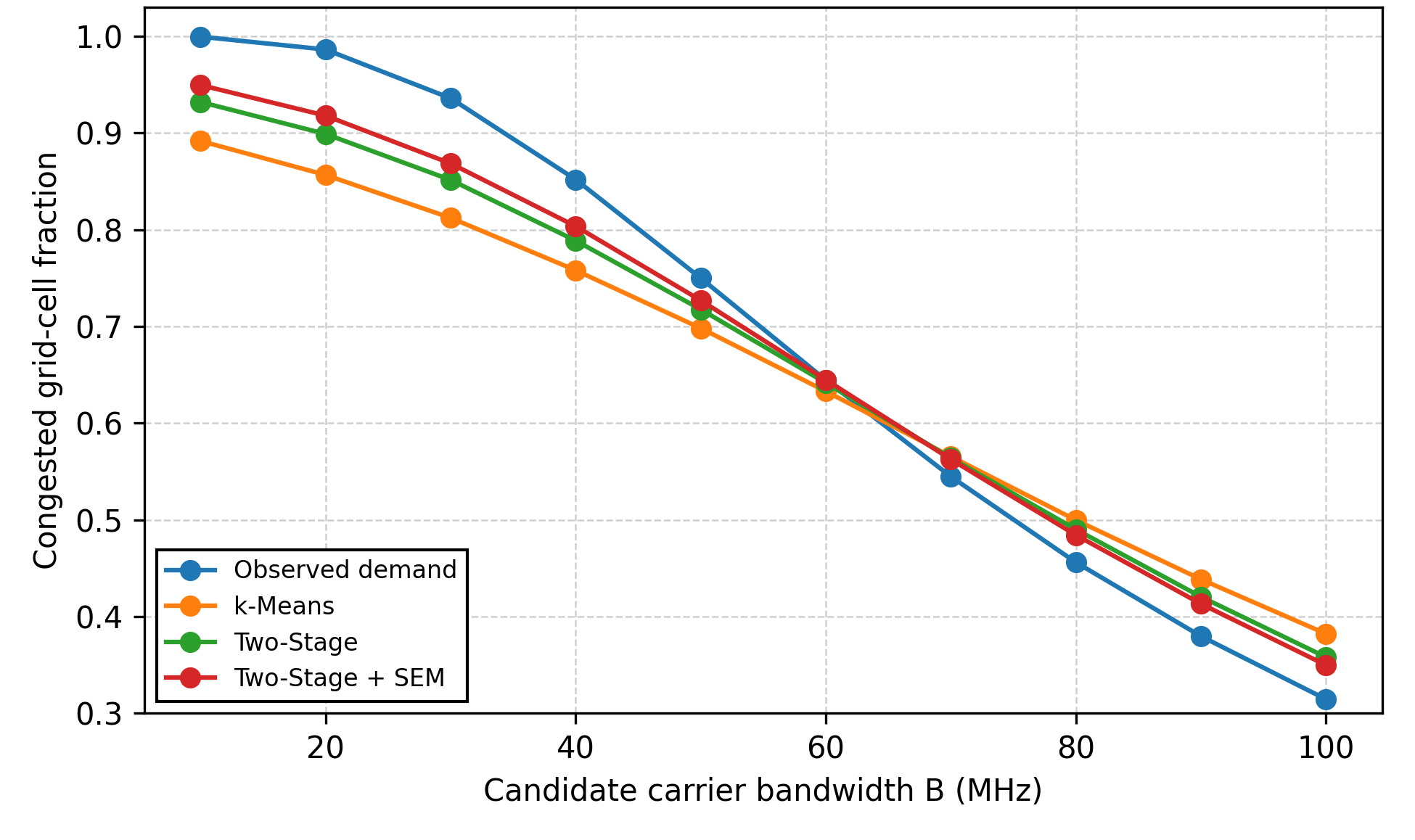}
    \caption{All Cities: $P_{\mathrm{cong}}(B)$ versus $B$ under observed and predicted demand.}
    \label{fig:cong}
\end{figure}

Figure~\ref{fig:cong} compares $P_{\mathrm{cong}}(B)$ under the reference surface $D_i$ (``Observed demand'') and predicted surfaces $\hat{D}_i$, where deviations from the reference quantify planning risk (overestimation/underestimation of congestion). Leakage-reduced splitting and SEM refinement improve spatial generalization and shift the inferred congestion curve toward the observed curve across $B$. For illustration without operator traffic traces, Fig.~\ref{fig:cong} uses an ``All Cities'' case study combining a heavy-tailed spatial demand distribution (log-normal proxy field) with prediction-error models calibrated to the MAE values in Table~\ref{tab:performance}, highlighting how moderate MAE differences translate into meaningful shifts in estimated congested area.

\section{Conclusion}\label{sec:conclusions}
This paper presented an AI-driven framework for spatial cellular traffic-demand prediction to support data-driven 5G/6G capacity and spectrum planning. The framework addresses spatial autocorrelation, which can inflate evaluation when training and validation sets are not properly separated, via a context-aware two-stage splitting strategy that reduces leakage while preserving functional representativeness across folds, and SEM refinement that mitigates residual spatial errors.

Evaluation across five major Canadian cities shows consistent MAE reductions relative to location-only clustering, with additional gains from SEM correction. A planning-oriented 5G NR mid-band case study demonstrates how the predictive gains translate into more reliable carrier bandwidth selection and provisioning assessments while controlling leakage and geographically coherent bias.

\bibliographystyle{IEEEtran}
\bibliography{biblio}

\end{document}